# Deep Labeling of fMRI Brain Networks

Ammar Latheef*[a], Sejal Ghate*[b], Zhipeng Hui[a], Alberto Santamaria-Pang[c], Ivan Tarapov[c], Haris I Sair[d,e], and Craig K Jones[a,d,e]

[a]Department of Computer Science, Johns Hopkins University, Baltimore MD 21218, USA
[b]Department of Biomedical Engineering, Johns Hopkins University, Baltimore MD 21218, USA
[c]Health AI, Microsoft, Redmond Washington
[d]Department of Radiology and Radiological Science, Johns Hopkins School of Medicine, Baltimore MD 08544, USA
[e]Malone Center for Engineering in Healthcare, Johns Hopkins University, Baltimore MD 21218, USA
*Co-first authorship

Corresponding Author:
Craig Jones, PhD
Malone Hall, Suite 340
3400 North Charles Street
Baltimore, MD 21218-2608
craigj@jhu.edu

## Abstract

Resting State Networks (RSNs) of the brain extracted from Resting State functional Magnetic Resonance Imaging (RS-fMRI) are used in the pre-surgical planning to guide the neurosurgeon. This is difficult, though, as expert knowledge is required to label each of the RSNs. There is a lack of efficient and standardized methods to be used in clinical workflows. Additionally, these methods need to be generalizable since the method needs to work well regardless of the acquisition technique. We propose an accurate, fast, and lightweight deep learning approach to label RSNs. Group Independent Component Analysis (ICA) was used to extract large scale functional connectivity patterns in the cohort and dual regression was used to back project them on individual subject RSNs. We compare a Multi-Layer Perceptron (MLP) based method with 2D and 3D Convolutional Neural Networks (CNNs) and find that the MLP is faster and more accurate. The MLP method performs as good or better than other works despite its compact size. We prove the generalizability of our method by showing that the MLP performs at 100% accuracy in the holdout dataset and 98.3% accuracy in three other sites' fMRI acquisitions.

## 1. Introduction

Clinical functional Magnetic Resonance Imaging (fMRI) has been utilized widely for several decades in the setting of preoperative brain mapping. More recently, resting state fMRI (rs-fMRI) has emerged as a viable tool to supplement task-fMRI, with evidence that many functional regions of the brain assessed by task-fMRI can be delineated with rs-fMRI. Furthermore, since rs-fMRI is based on connectivity patterns of the whole brain, multiple intrinsic brain networks (IBNs) can be elucidated using a single acquisition, conferring benefit over task-fMRI which requires a specific subset of brain function to be interrogated. Rs-fMRI indeed has been shown to be useful in the setting of preoperative brain mapping (Lee et al., 2013; Seitzman et al., 2019).

Labelling of RSNs requires expertise and can be time consuming (Roquet et al., 2016). Despite its potential uses in the clinical setting, there is no standard metric or method for labelling of fMRI RSNs for use in clinical workflows, and frequently in the literature, similar appearing brain networks are labeled as different networks. When performing hypothesis driven analysis, for example seed or region of interest (ROI) based analysis, the resultant IBN can be labeled as part of the system related to the functional specialization of region of interest. Independent component analysis (ICA) (Beckmann, 2012), the other major method of rs-fMRI analysis, does not require such an a priori approach (Lv et al., 2018), and offers benefit in situations where issues like brain shift (often present in clinical populations undergoing fMRI for preoperative brain mapping) or nonstandard anatomy can compromise seed/ROI placement.

Previous methods for automatic ICA labelling used template matching (Greicius et al., 2004; Coulborn et al., 2021). In template matching, a similarity score is calculated between a reference image for each ICA label and the target image. More recent methods use deep learning. Deep learning on fMRI has been widely used in neuroscience studies (Wen et al., 2018). Nouzet et al. (2020) found that MLPs can be used for the classification of rs fMRI ICA. Chou et al. (2018) and Zhao et al. (2018) applied other deep learning methods on rs-fMRI ICA and achieved high accuracies. 3D CNNs have been shown to be good in identifying features in spatial data. Chou et al. (2015) used a deep Siamese network to classify rs-fMRI ICA components with high accuracy as well as for one-shot generalization. Luckett et al. (2020) found that a 3DCNN can be used with rs-fMRI for accurately and specifically localizing the language network in patients with brain tumors.

To extend this prior work, we propose 1) a highly accurate model that is small and can be reliably run through a reproducible pipeline on the cloud, 2) that supports a larger number of RSNs than previous methods, and 3) is generalizable across unseen cohorts. We hypothesize that a deep learning model can accurately classify brain networks generated from ICA.

## 2. Materials and Methods

### 2.1. Dataset

To demonstrate feasibility and reproducibility we used data from the publicly available 1000 Functional Connectomes Project (Biswal et al., 2010). We analyzed four cohorts: i) Beijing Zhang (N=198 [76M/122F]; ages: 18-26; number slices = 33; number timepoints = 225) ii) Bangor (N=20 [20M/0F]; ages: 19-38; number slices = 34; number timepoints = 265), iii) New York, (N=84 [43M/41F]; ages: 7-49; number slices = 39; number timepoints = 192), and iv) Palo Alto (N=17 [2M/15F]; ages: 22-46; number slices = 29; number timepoints = 235). The TR was 2s for all cohorts. Other imaging parameters are found at the 1000 Functional Connectomes Project site.

### 2.2. Processing of Data

2.2.1 Image Preprocessing
All fMRI data was motion corrected, spatial smoothed, temporal high pass filtered, and registered into MNI space using the FMRIB Software Library v6.0 (FSL) (Smith et al., 2004; Jenkinson et al., 2012). Default parameters were used for motion correction and temporal filter and a FWHM = 7mm was used for spatial smoothing.

2.2.2 Independent Component Analysis (ICA) and Dual Regression

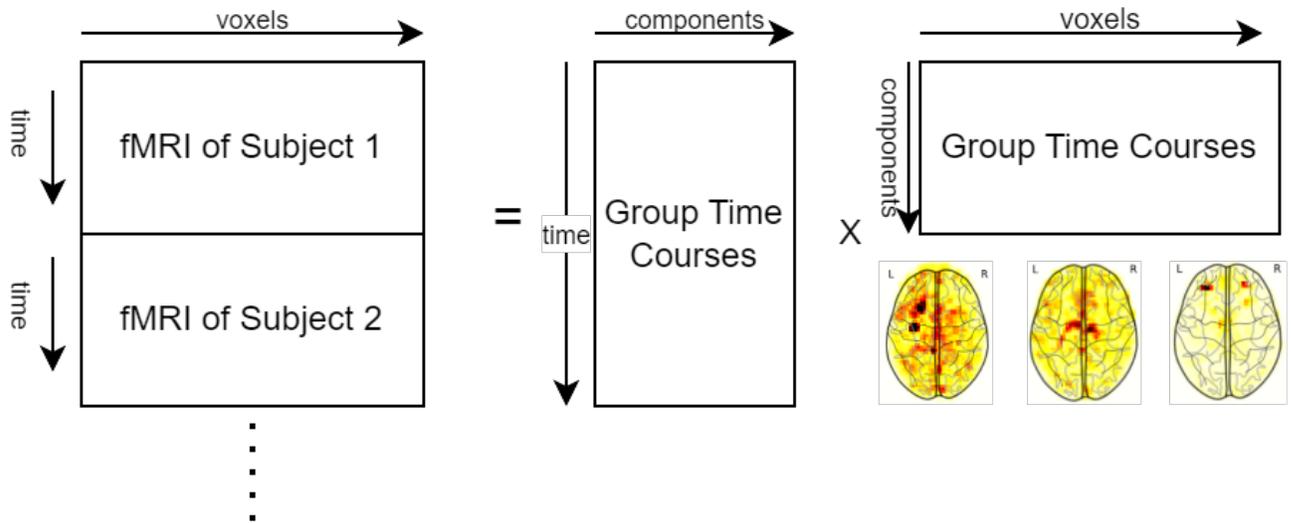

**Fig. 1** Group-ICA maps are created from a concatenated matrix of fMRI data and then solved for group time courses and group ICA maps

Group ICA (Abou Elseoud et al., 2011) was performed on the Beijing Zhang dataset using MELODIC (Beckmann and Smith, 2004) in FSL with target components spanning 40, 60, 80, and 100 components. 100 component ICA maps were used for subsequent analyses. Dual regression was then performed using the group ICA maps as the templates to back-project to each subject's fMRI data; time courses from these back-projected maps were then used as regressors to create subject specific ICA maps. The order of specific ICA components thus is consistent between the group ICA maps and the subject specific ICA maps – this enables efficient labeling of a large sample of subject specific ICA maps.

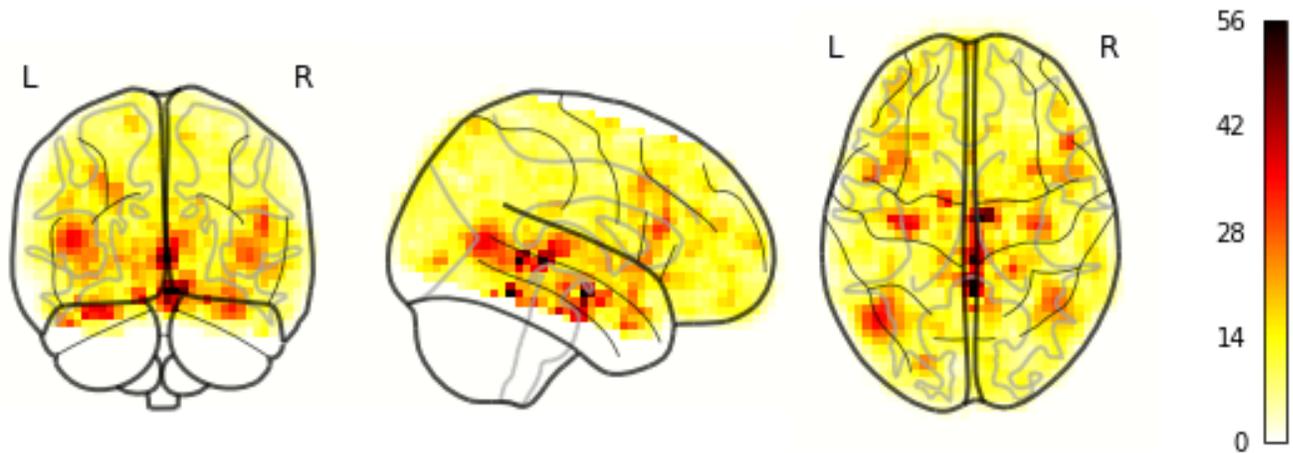

**Fig. 2** Example dual regression activation map of the visual lateral superior class

Quality Control: After visual checking of data, 22, 4, and 13 samples from Bejing-Zhang, New York, and Bangor cohorts, respectively, were found to be misregistered and subsequently discarded. We did not discard any sample from the Palo Alto cohort.

2.2.3 Labeling of group ICA maps

Manual labeling of the ground truth data was done by a neuroradiologist with 12 years of resting state fMRI experience on the group ICA maps; the dual regression process identified subject-level maps to which the group ICA label maps were applied. Labels were generated for each component based on the topology of the network. Labels were generated at the highest level of granularity based on the specific subsystem of a larger networks – for example, the motor network can be broken down into left and right, as well as dorsal and ventral motor system, and the dorsal motor system can be broken down into upper extremity (in our case, localizing to the "omega" of the precentral gyrus indicating the hand motor region), and lower extremity (involving the more medial paracentral lobule). Lower order ICA maps were used as a guide to confirm relationships between the networks at 100 components. Other works have also shown the brain networks as hierarchically organized (Doucet et al., 2011; Vidaurre et al., 2017). ICA component time courses and power spectra were utilized to guide distinction between noise and real network components. We identified a total of 58 unique RSN labels across the dataset, including "noise" and "unknown". The noise label was used when the ICA map did not look like a real component and instead looked like random noise, including signal outside of the brain, in the ventricular system, related to vascular signal, and in the white matter. The unknown label was used when the ICA map looked like a real component, but it was not possible to attribute it to a specific IBN. (See Appendix for the list of RSN labels.)

## 2.3. Neural Networks Comparison

Three neural network architectures were compared: 1) a Multilayer Perceptron (MLP), 2) a Multi-Projection CNN (MP-CNN), and 3) a fully 3D Convolutional Neural Networks (3D CNN). Each is described in further detail below. The dataset was randomly split into three sets: 72% for training (n=126), 8% for validation (n=14), and 20% for testing (n=40). The fMRI time series from each subject had been decomposed into 100 components, thus, the number of training samples was 12,600. The validation dataset was 1400 samples, and the testing dataset was 4000 samples. Each component was represented as a matrix of size 45 x 54 x 45 and was classified into one of 58 RSN labels. We used the SGD optimizer, learning rate of 1e-5, and cross entropy loss with labels weighted according to their distribution in the dataset. The trained networks were tested on the holdout set of Beijing Zang and three other cohorts: Palo Alto, Bangor, and New York. All training was done in Azure ML using an NVIDIA Tesla K80 GPU compute cluster. The neural network structures are further described below.

### 2.3.1 Multilayer Perceptron (MLP)

The MLP consisted of three hidden layers of 200 nodes each and a dropout of 66% after the second layer. Input data was 3D dual regression activation maps flattened into a 1D array. A further ablation study was performed to determine the smallest number of layers and nodes needed for equivalent accuracy. The final output was a vector of probabilities of size 58 (equal to the number of labels).

### 2.3.2 Multi-Projection CNN (MP-CNN)

A 2.5D multi-projection CNN network was created based on a pre-trained 2D Resnet-50 (all layers unfrozen). The input data was created by combining axial, sagittal and coronal projections into an RGB representation where the axial projection was coded as the red channel, sagittal projection as the green channel, and coronal projection as the blue channel. Each projection was a maximum signal intensity projection along the axis. The final output was a vector of probabilities of size 58 (equal to the number of labels).

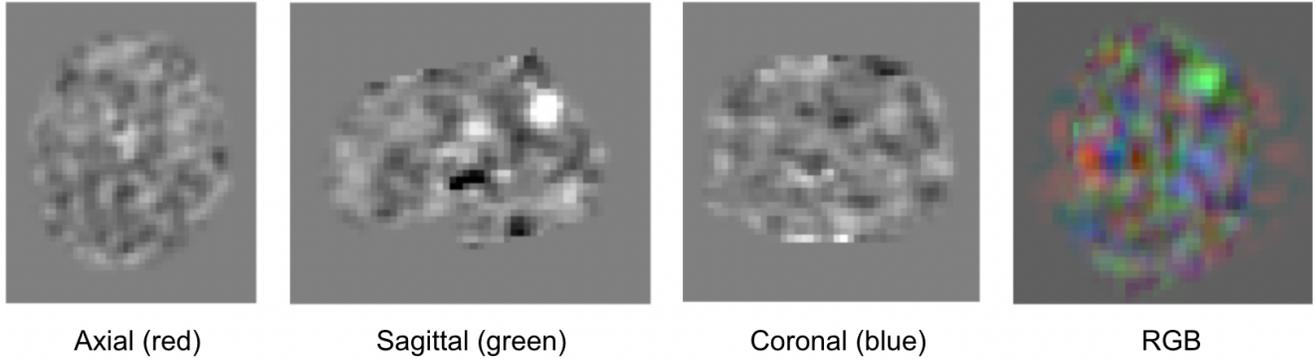

| Axial (red) | Sagittal (green) | Coronal (blue) | RGB |

**Fig. 3** RGB Image construction for input to the MP-CNN Network. ICA output was projected into the axial plane (left), sagittal plane (middle-left), and coronal plane (middle-right) as maximum intensity projections. A 2D RGB image was created from the three projections by using the axial projection in the red channel, the sagittal project in the green channel, and the coronal project in the blue channel

### 2.3.3 3D CNN

A 3D CNN network was created based on the pre-trained 3D Resnet (r3d_18 in PyTorch; all layers unfrozen). The 3D CNN has been used in other works that analyze fMRI ICA (Nouzet et al., 2020; Chou et al., 2018; Zhao et al., 2018). The 3D CNN input was the subject specific ICA maps without any flattening or projection of the data. The output was a vector of SoftMax prediction values of size 58 as in the other two neural networks.

### 2.4. MLP Ablation Study

The MLP ablation study was applied on Beijing Zhang's dataset where a stratified 5-fold cross-validation scheme was used to compare the accuracies between different classifiers. To search for the minimum values of the number of units of the MLP architecture with great performance, the grid search was deployed with:

Number of Layers: [1, 2, 3]

Number of nodes per layer: [2, 5, 10, 20, 50, 100, 150, 200]

For each classifier, the accuracy is the mean value of 5 folds' validation sets' accuracies. All classifiers would stop training when threshold of loss reaches 0.0005 which reduced the overfitting possibility. The activation function between hidden layers was ReLU. The loss function was categorical cross-entropy, and the optimizer was SGD. The output was modified by Softmax activation function to map the 58 clusters into confidence distribution. To be clear, the learning rate and dropout value is fixed as 1e-5 and 0 since the underfitting issue is detected in bigger values of them. Due to the imbalanced distribution of clusters (most of the clusters' labels are noise), the class weight method was used.

# 3. Results

## 3.1. Training Results

Each neural network was trained on the training portion of the Beijing Zhang dataset. Final training accuracies were 99.8%, 99.4%, and 98.1% on MLP, MP-CNN, and 3D CNN respectively.

## 3.2. Test results

The three trained neural networks were evaluated on the hold out test set of the Beijing Zhang data and the other three cohorts, none of which were used for training. As shown in **Table 1** all three neural network architectures performed well on the test for generalization, with MLP being the most accurate and the fastest prediction times. The MLP achieved 100% accuracy on the Beijing Zhang test holdout and the Palo Alto cohort. The MP-CNN had a slightly higher mean accuracy across datasets than the 3D-CNN but was still worse than the MLP. It performed well on the hold out test portion of the Beijing Zhang cohort, the same cohort that it was trained on. The 3D CNN performed well on all cohorts except Bangor, which only has 7 participants (700 activation maps) to classify. MLP ran the fastest among the models – it predicted results approximately twice as fast as the MP-CNN and 40 times faster than the 3D CNN model.

|        | Beijing (n=36) | New York (n=80) | Palo Alto (n=17) | Bangor (n=7) |
|--------|----------------|-----------------|------------------|--------------|
| MLP    | 100.0          | 99.94%          | 100%             | 98.3%        |
| MP-CNN | 99.5           | 97.5            | 91.7             | 93.6         |
| 3D CNN | 98.8           | 97.6            | 95.5             | 91.1         |

**Table 1** Mean prediction accuracies (percentage) for each model during testing. For the Beijing-Zhang dataset, 36 of 176 participants were used for testing

## 3.3. Misclassification

Misclassified data were manually reviewed. The majority of the misclassifications occurred between real network components and noise or unknown, as the spatial distribution of some noise and unknown components may have overlapped that of a real network. The other less common error was due to adjacent or slightly overlapping anatomical locations of subcomponents of different brain networks.

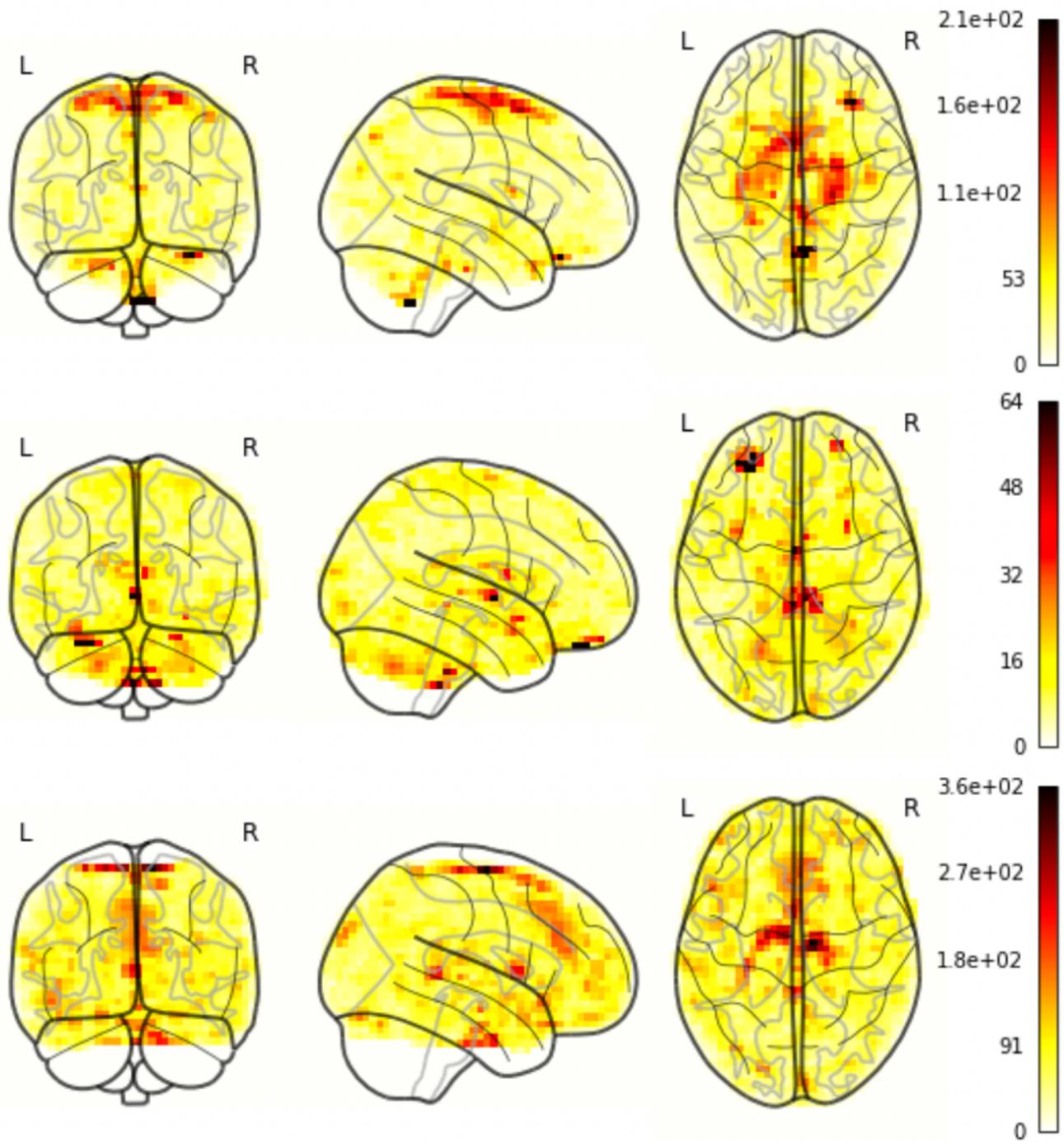

**Fig. 4** 3 different example misclassifications by our models. From top to bottom: a) misclassified noise as attention-fef, top 3 predictions: attention-fef: 0.446, motor-dorsal-leg: 0.237, noise: 0.065, b) misclassified noise as visual-lingual-posterior, top 3 predictions: visual-lingual-posterior: 0.185, noise: 0.129, visual-lateral-inferior-left: 0.113, c) misclassified noise as cerebellar-lateral-right, top 3 predictions: cerebellar-lateral-right: 0.257, cerebellar-superior-mid-posterior: 0.245, noise: 0.197

**Fig. 5** shows the confusion matrix on the predictions of the worst performing model (3D model on Palo Alto cohort). Most regions have a high probability of being predicted correctly. The exception is among Noise, Unknown, and Cerebellar components, where the images have a relatively high probability to be misclassified.

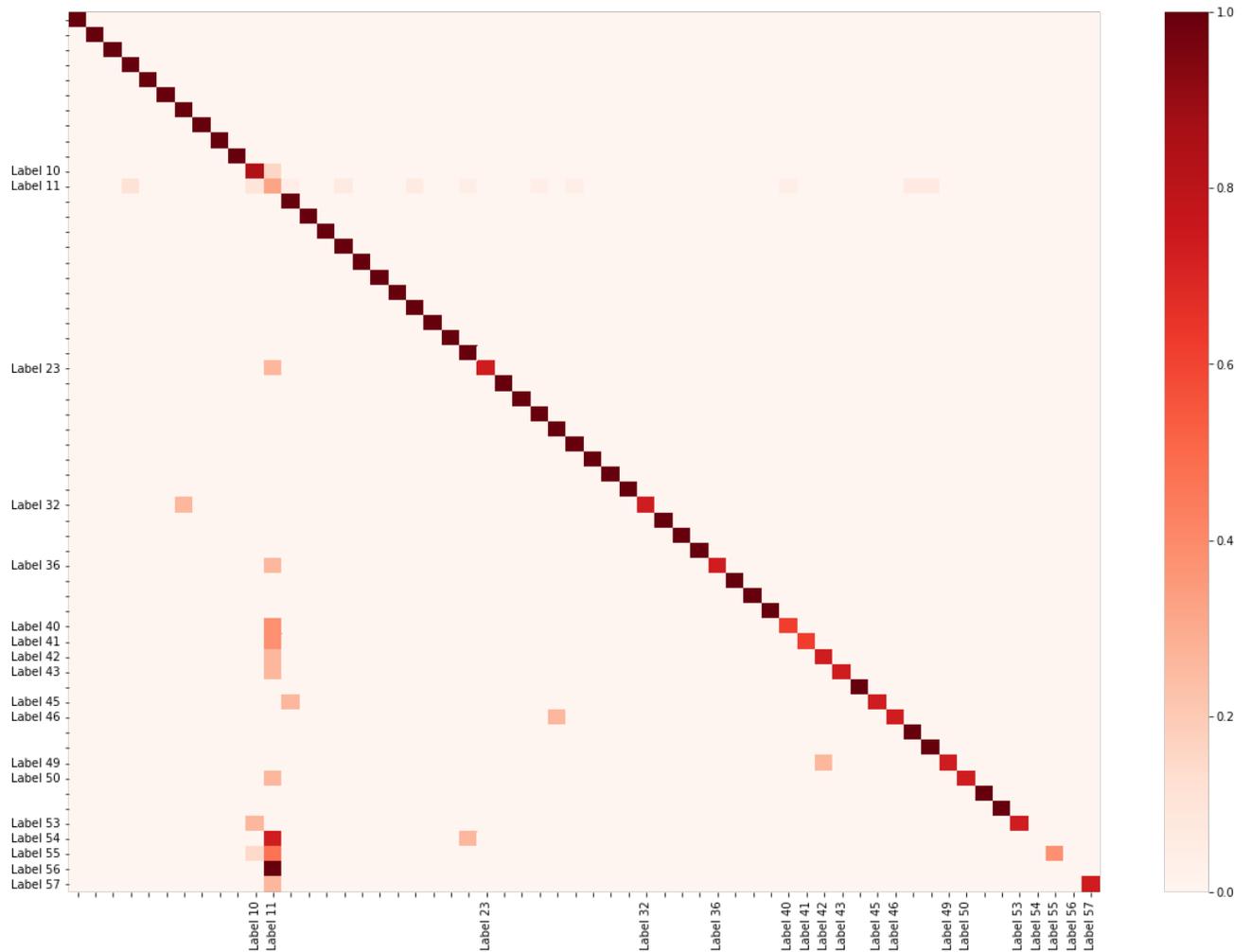

**Fig. 5** Normalized confusion matrix heatmap with logarithmic scaling, where the color or shading of each cell indicates the relative magnitude of the corresponding value in the confusion matrix, and the rows represent the true labels. Only regions with at least one corresponding misclassification have been labeled

## 3.4. MLP Ablation Study

In **Fig. 7**, the accuracy of the model increases rapidly with the number of nodes, reaching 98% at 10 nodes and a peak of almost 100% at 20 nodes and 1 hidden layer. After that, the accuracy remains high and does not significantly change. Figure 7 shows the mean prediction accuracies of cases with number of nodes [2, 5, 10, 20, 50, 100, 150, 200] and number of hidden layers [1,2,3].

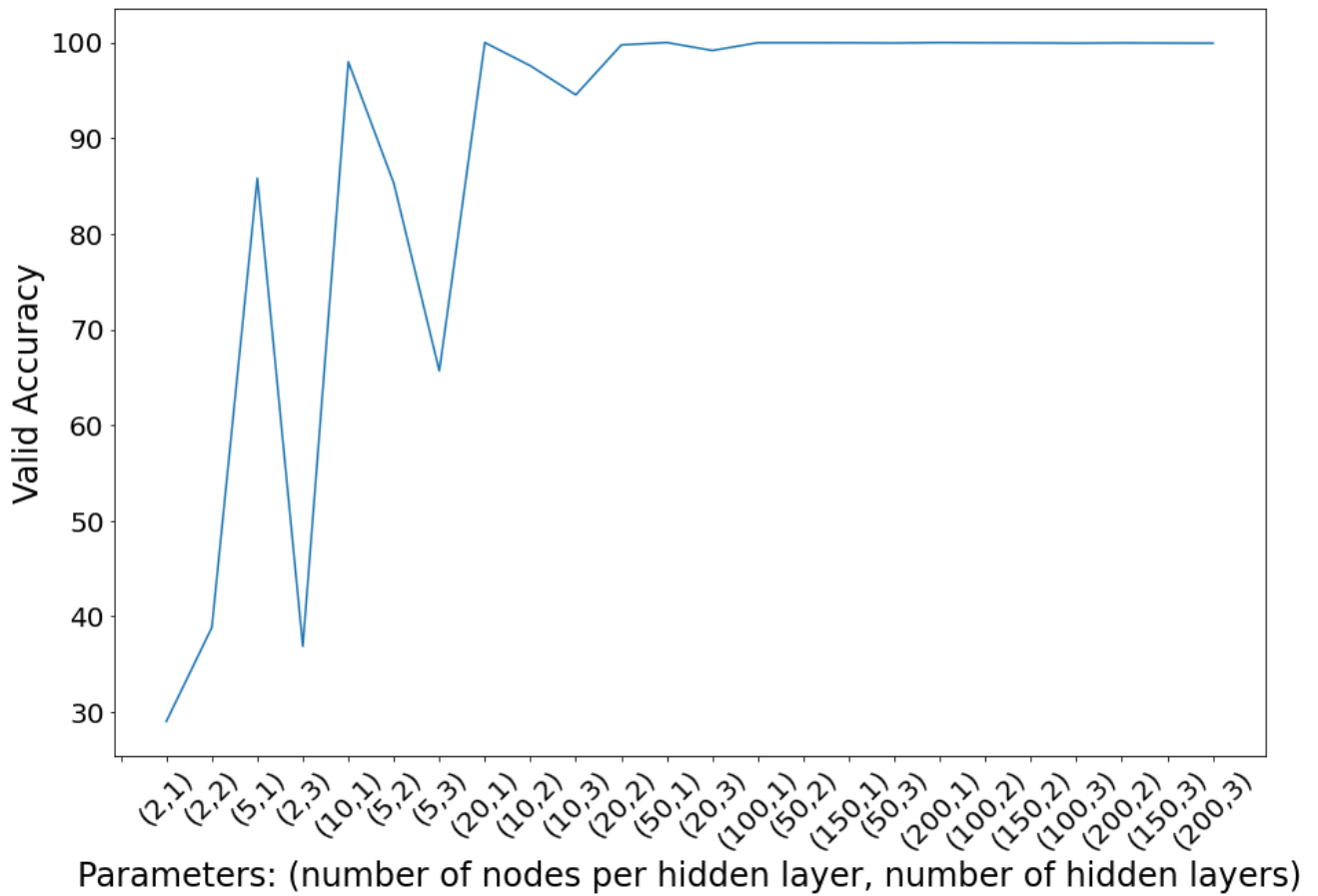

**Fig. 6** Mean prediction accuracy across different number of layers and nodes per layer. The x axis is sorted by the total number of nodes.

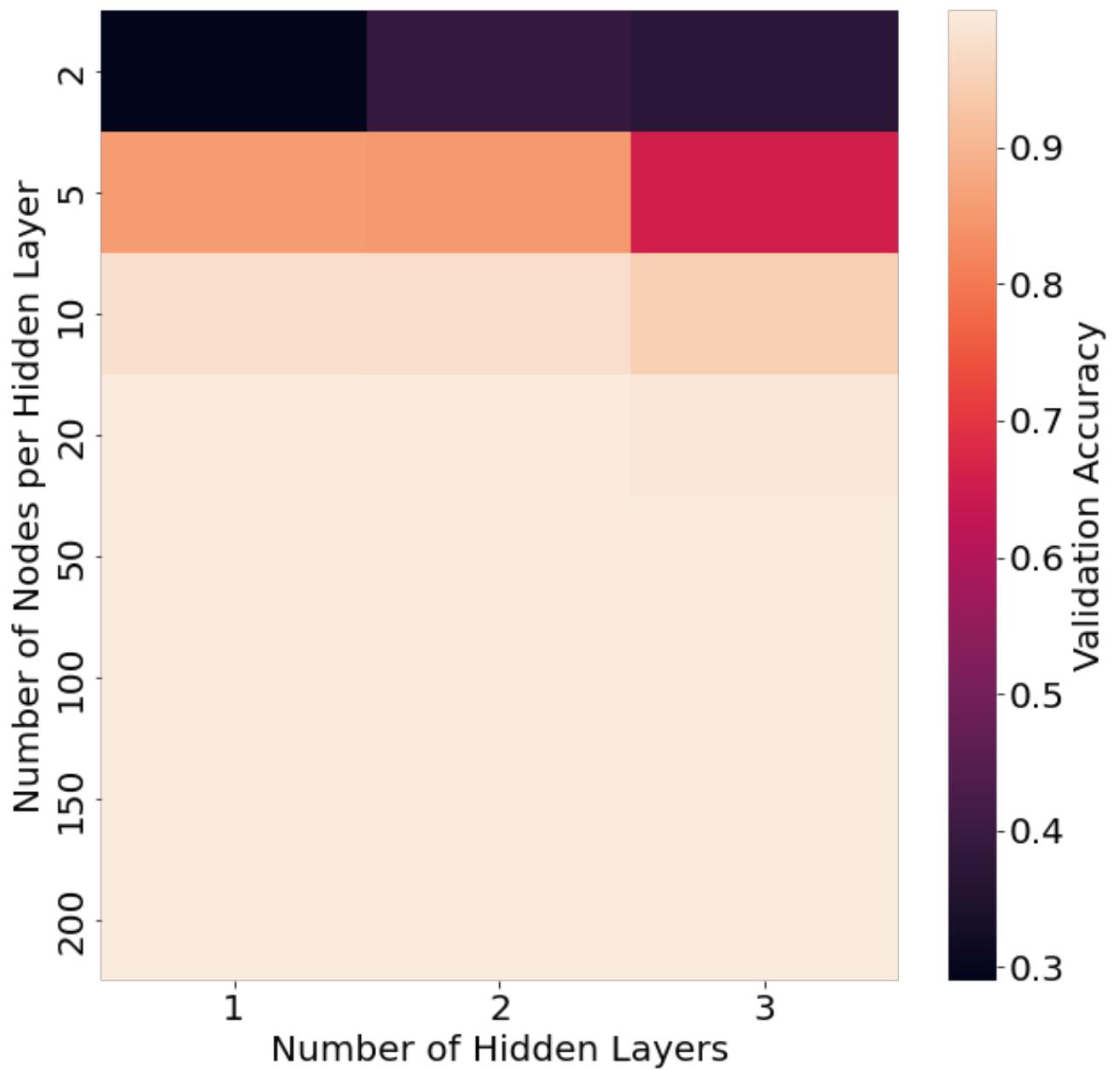

**Fig. 7** Heatmap of mean prediction accuracies for across different number of layers and nodes per layer. Each cell represents the mean accuracy of 5-fold validation sets.

## 3.5. Time for prediction and training

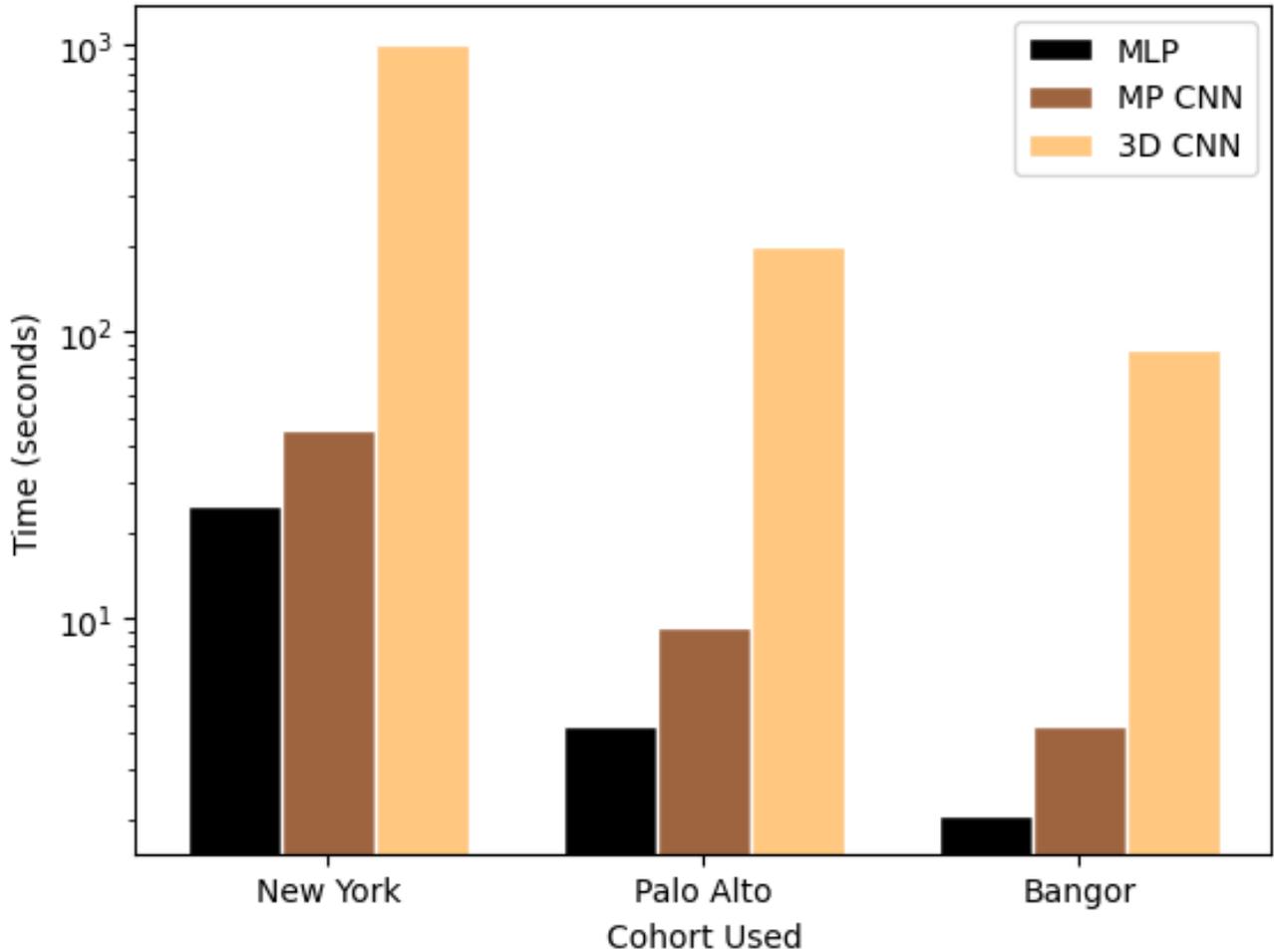

**Fig. 8** Total inference time taken for the cohorts used in our models on a logarithmic time scale

Our 3D CNN took a large amount of time during inference, so we also tried a 2D CNN by projecting our 3D activation maps on a 2D plane. The 2D CNN performed significantly worse than both the 3D CNN and MLP models. While it runs faster than the 3D CNN, it is slower than the MLP, making it undesirable in comparison, as can be seen in Figure 8. Our MLP, on the other hand, was very fast during inference, running twice as fast as the MP CNN and 40 times as fast as the 3D CNN on average. The total inference time for each of the cohorts simply depended on the number of patients.

# 4. Discussion

## 4.1. Summary

We ran the models on different architectures to find the model that performed best considering the accuracy on unseen data, inference time, and model size. The MLP network performed the best on metrics of accuracy (greater than 98% accuracy across all datasets), and speed (inference of about 330 samples/second). It also is the most compact of the three models. On running ablation experiments with smaller MLPs, we found that our MLP needs only a small hidden layer to perform highly accurately.

## 4.2. Comparison to Others

Nouzet et al. (2020) Reached an accuracy of 92 % with MLPs. They tried running with many configurations of MLP to find the most suitable parameters. The best configuration using a grid search was found to be 3 layers of 5,120 nodes, wi8th rectified linear unit (ReLU) activation, a 0.66 dropout rate and a learning rate of 10e-5. Our MLP achieved over 98.3% accuracy across three unseen cohorts, which is much higher than other studies (Nouzet et al., 2020; Chou et al., 2018; Zhao et al., 2018). Our MLP was also much smaller and had to classify from a larger number of RSNs. For instance, Nouzet et al. (2020) considered 45 RSNs and Chou et al. (2018) considered eight.

Our results showed that MLPs perform better than CNNs. This is in line with Nouzet et al. (2020), who also selected MLPs over CNNs because they found that the small datasets were problematic for the large CNN models leading to overfitting.

One potential factor is the difference in preprocessing techniques. We performed Group ICA, creating a shared set of independent components for all participants. We then back projected it onto the individual maps, aligning each participant's map onto the global components. In contrast, others used ICA maps generated on individual subjects, leading to independent components of one subject that do not necessarily align with others. Thus, our final ICA maps might be easier to classify for deep learning networks due to the consistent alignment of the independent components across all participants. Another possible reason for lower accuracy in Nouzet et al. (2020) is that they created ground truth labels automatically using an MLP. Chou et al. (2018) reported an accuracy of 98% with CNNs, which is comparable to ours on the original dataset (Beijing cohort). But their best performing model, the CNN, was much larger than our MLP in the number of parameters and they only used eight RSNs.

## 4.3. Misclassifications

Almost all misclassifications, which were infrequent, involved noise or unknown, i.e., noise was misclassified as a network, or a network got misclassified as noise. For instance, across the 73 misclassifications in the Palo Alto cohort with the 3D model, only one misclassification did not have both noise and unknown (Misclassified attention-dorsal-anterior-right as cerebellar-lateral-right)

Another source of error is dual regression. Dual regression may incorrectly back-project the ICA map on the individual subject. For example, in Figure 9, executive-anterior-ventral has been projected onto a subject incorrectly, with a large gap in the anterior part. The resulting activation map does not belong to the executive-anterior-ventral category. Our ground truth labelling was done on the group ICA maps, instead of each dual regression map. This is because labelling each dual regression map would be highly time consuming. Another reason for not labelling at the subject level is because this is another source of labelling error since each is labelled separately.

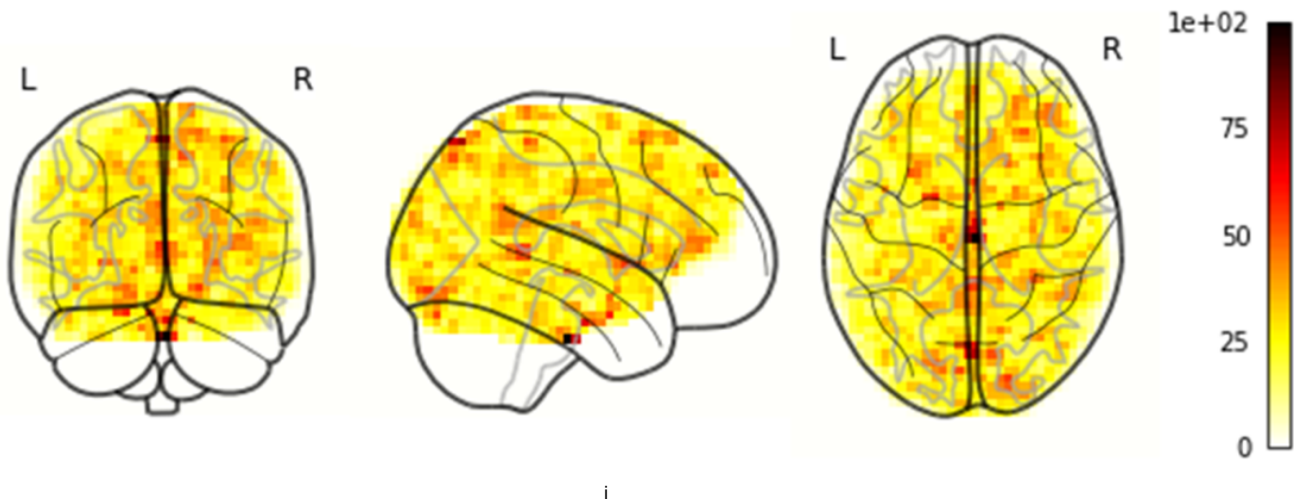

i

**Fig. 9** Dual regression has stripped the anterior part of the activation map during back-projection

In many of the misclassifications, the 2$^{nd}$ or 3$^{rd}$ potential labels defined by softmax predictions represented the true label as can be seen in figure 4.

Misclassification of noise:

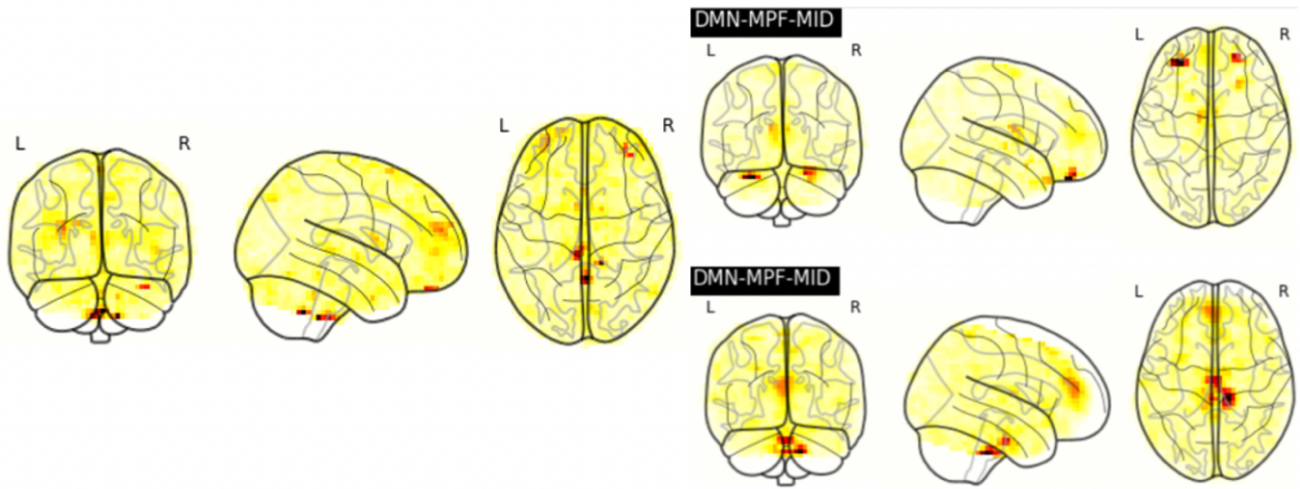

**Fig. 10** Clockwise, starting from the left a) Noise RSN that was misclassified as DMN-MPF-MID, b) a sample DMN-MPF-MID network for comparison, and c) another sample DMN-MPF-MID network

In some misclassifications, the misclassified networks have some similar characteristics to the network classified by the model. In the sample misclassification on figure 10, the noise map has strong signal in the anterior and the inferior regions which might have fooled the model into misclassifying this as DMN-MPF-MID.

### 4.4. Limitations and Future Work

The method of utilizing dual regression allowed for efficient labeling of a high number of data samples. Individual samples may have incorrect labels based on dual regression. However, we postulate that a large number of incorrectly labeled data would not allow creation of a model that generated the highly accurate results we demonstrated in the various test datasets. A method to determine the similarity of the dual regressed ICA maps and the group ICA map, or a metric of confidence of the subject specific map, may be considered or developed in the future. Alternatively, label error detection algorithms may be utilized to identify potentially mislabeled individual data.

We also focused our model development using 100 target ICA components, which is somewhat of an arbitrary choice. With 100 target components, the aim was to balance the depiction of small subcomponents of major intrinsic brain networks, versus merging of different networks into one component. For each subject, the ideal target number may be different to optimally separate networks. We used lower order ICA to help identify networks in the 100 ICA dataset. Future studies may consider incorporating each of the different ICA orders and utilizing a hierarchical learning method to further improve accuracy of network detection across multiple scales.

We limited datasets to those with TR of 2 seconds. This was done partly to ensure that all the data included was able to be used concurrently as part of group ICA. Another reason was that TR of 2 seconds is widely used for clinical fMRI. Higher sampling rates are used frequently in fMRI research now, and in some clinical settings. Evaluation of methods across different fMRI sampling rates may be performed in the future to ensure further generalizability.

## 5. Conclusion

Deep learning can be used for classification of fMRI labels with good generalizability across fMRI datasets. The MLP model performed best, implying that relatively simple model architectures can capture the features of resting state networks from activation maps. Our MLP model is small, quick, and accurate. For reproducibility, we deployed it on Microsoft Azure Machine Learning cloud platform. Considering all these factors, our pretrained MLP can be used off-the-shelf in the clinical setting.

## 6. Information Sharing Statement

The data was obtained from the publicly available 1000 Functional Connectomes Project (http://fcon_1000.projects.nitrc.org/fcpClassic/FcpTable.html). We used the Beijing, New York, Palo Alto, and Bangor cohorts. The FSL software used in this study is available for download (https://fsl.fmrib.ox.ac.uk/fsl/fslwiki/FslInstallation).

## 7. Declarations

### 7.1. Funding

This research did not receive any specific grant from funding agencies in the public, commercial, or not-for-profit sectors.

### 7.2. Competing Interests

The authors have no competing interests to declare that are relevant to the content of this article.

## 7.3. Ethics Statement

The publicly available 1000 Functional Connectomes Project was collected in accordance with HIPAA guidelines. It is anonymous and contains no protected health information.

- Zhao, Y., Dong, Q., Zhang, S., Zhang, W., Chen, H., Jiang, X., Guo, L., Hu, X., Han, J., & Liu, T. (2018). Automatic recognition of fmri-derived functional networks using 3-d convolutional neural networks. *IEEE Transactions on Biomedical Engineering*, *65*(9), 1975–1984. https://doi.org/10.1109/TBME.2017.2715281

# 9. Appendix

1) Reproducibility
2) RSN Labels

## 9.1. Reproducibility

- All our experiments were done on Azure ML.
- We created a pipeline for reproducibility.
- Further, the dataset we used is publicly available from the 1000 Functional Connectomes Project.

## 9.2. RSN Labels

Labeling structure: Function – Location – Sub-location

List of Functions: DMN, Attention, Visual, Motor, Executive, Salient, Sensory, Basal Ganglia, Cerebellar, Visual-Lateral. Auditory, Cognitive.

Locations: Lateral, PCC, RSC, Ventral, Lingual, Posterior, Dorsal, MPF, ACC, Anterior, mid, right, left, medial, leg, caudate.

The list of labels used are

| | | |
|---|---|---|
| 1 | 0 | DMN-RSC |
| 2 | 1 | DMN-PCC-MID |
| 3 | 2 | DMN-MPF-VENTRAL |
| 4 | 3 | DMN-MPF-MID |
| 5 | 4 | ATTENTION-DORSAL-IPS-MID |
| 6 | 5 | MOTOR-VENTRAL-SUBCENTRAL |
| 7 | 6 | VISUAL-LINGUAL-ANTERIOR |
| 8  | 7  | ATTENTION-DORSAL-IPS-MEDIAL |
| 9  | 8  | DMN-PCC-POSTERIOR |
| 10 | 9  | DMN-PCC-ANTERIOR |
| 11 | 10 | UNKNOWN |
| 12 | 11 | NOISE |
| 13 | 12 | ATTENTION-DORSAL-IPS-LATERAL-RIGHT |
| 14 | 13 | MOTOR-VENTRAL |
| 15 | 14 | EXECUTIVE-POSTERIOR-LEFT |
| 16 | 15 | NOISE |
| 17 | 16 | MOTOR-DORSAL-LEG |
| 18 | 17 | SENSORY-DORSAL-HAND-RIGHT |
| 19 | 18 | NOISE |
| 20 | 19 | VISUAL-CUNEUS-SUPERIOR-LATERAL-POSTERIOR |
| 21 | 20 | EXECUTIVE-POSTERIOR-RIGHT |
| 22 | 21 | VISUAL-CUNEUS-MEDIAL |
| 23 | 22 | SENSORY-DORSAL-HAND-LEFT |
| 24 | 23 | DMN-MPF-DORSAL |
| 25 | 24 | NOISE |
| 26 | 25 | NOISE |
| 27 | 26 | NOISE |
| 28 | 27 | SENSORY-DORSAL-MEDIAL |
| 29 | 28 | VISUAL-FUSIFORM-POSTERIOR |
| 30 | 29 | SALIENCE-INSULA-ANTERIOR |
| 31 | 30 | SALIENCE-ACC-MID |
| 32 | 31 | NOISE |
| 33 | 32 | ATTENTION-FEF |
| 34 | 33 | ATTENTION-DORSAL-IPS-LATERAL-LEFT |
| 35 | 34 | DMN-CINGULATE-MID |
| 36 | 35 | NOISE |
| 37 | 36 | AUDITORY-RIGHT |
| 38 | 37 | NOISE |
| 39 | 38 | DMN-IPL-LEFT |
| 40 | 39 | NOISE |
| 41 | 40 | COGNITIVE-MFG |
| 42 | 41 | BASALGANGLIA-LENTIFORM |
| 43 | 42 | LANG-BROCA |
| 44 | 43 | SALIENCE-INSULA-POSTERIOR |
| 45 | 44 | NOISE |
| 46 | 45 | NOISE |
| 47 | 46 | UNKNOWN |
| 48 | 47 | ATTENTION-DORSAL-ANTERIOR-RIGHT |
| 49 | 48 | SALIENCE-ACC-ANTERIOR |
| 50 | 49 | NOISE |
| 51 | 50 | EXECUTIVE-ANTERIOR-VENTRAL |
| 52 | 51 | NOISE |
| 53 | 52 | MOTOR-DORSAL-LEG |
| 54 | 53 | DMN-IPL-RIGHT |
| 55 | 54 | NOISE |
| 56 | 55 | VISUAL-CUNEUS-SUPERIOR-LATERAL-ANTERIOR-LEFT |
| 57 | 56 | DMN-HIPPOCAMPUS |
| 58 | 57 | DMN-PARAHIPPOCAMPAL |
| 59 | 58 | CEREBELLAR-SUPERIOR-MEDIAL |
| 60 | 59 | NOISE |
| 61 | 60 | BASALGANGLIA-VENTRAL-NA |
| 62 | 61 | AUDITORY-LEFT |
| 63 | 62 | MOTOR-DORSAL-HAND |
| 64 | 63 | LANGUAGE-WERNICKE |
| 65 | 64 | UNKNOWN |
| 66 | 65 | THALAMUS-MEDIAL |
| 67 | 66 | NOISE |
| 68 | 67 | UNKNOWN |
| 69 | 68 | VISUAL-LINGUAL-POSTERIOR |
| 70 | 69 | UNKNOWN |
| 71 | 70 | NOISE |
| 72 | 71 | VISUAL-FUSIFORM-ANTERIOR |
| 73 | 72 | UNKNOWN |
| 74 | 73 | NOISE |
| 75 | 74 | VISUAL-LATERAL-INFERIOR-LEFT |
| 76 | 75 | NOISE |
| 77 | 76 | VISUAL-LATERAL-SUPERIOR |
| 78 | 77 | NOISE |
| 79 | 78 | NOISE |
| 80 | 79 | NOISE |
| 81 | 80 | NOISE |
| 82 | 81 | NOISE |
| 83 | 82 | NOISE |
| 84 | 83 | NOISE |
| 85 | 84 | NOISE |
| 86 | 85 | NOISE |
| 87 | 86 | BASALGANGLIA-CAUDATE |
| 88 | 87 | VISUAL-LATERAL-INFERIOR-RIGHT |
| 89 | 88 | CEREBELLAR-SUPERIOR-MID-POSTERIOR |
| 90 | 89 | NOISE |
| 91 | 90 | NOISE |
| 92 | 91 | CEREBELLAR-LATERAL-RIGHT |
| 93 | 92 | NOISE |
| 94 | 93 | CEREBELLAR-SUPERIOR-MID-ANTERIOR |
| 95 | 94 | NOISE |
| 96 | 95 | CEREBELLAR-LATERAL-RIGHT |
| 97 | 96 | NOISE |
| 98 | 97 | NOISE |
| 99 | 98 | HYPOTHALAMUS |
| 100 | 99 | NOISE |